\documentclass[3p]{elsarticle}

\usepackage{array}
\newcolumntype{P}[1]{>{\centering\arraybackslash}p{#1}}

\makeatletter
\newcommand\footnoteref[1]{\protected@xdef\@thefnmark{\ref{#1}}\@footnotemark}
\makeatother

\usepackage{amssymb}
\usepackage{pifont}

\newcommand{\xmark}{\ding{55}}

\usepackage[retain-explicit-plus]{siunitx}
\usepackage{lineno,hyperref}
\usepackage{mathtools}
\usepackage{multirow}
\usepackage{multicol}
\usepackage{booktabs}
\usepackage{listings}
\usepackage{makecell}
\usepackage{tablefootnote}
\usepackage{subcaption}
\usepackage{threeparttable}
\usepackage{xcolor}
\usepackage{footmisc}
\usepackage[font=normalsize]{caption}
\usepackage{algorithmic}
\usepackage[ruled,vlined]{algorithm2e}
\journal{Entertainment Computing}
\usepackage{blindtext}
\usepackage{hyperref}
\usepackage{nameref}

\newcounter{mylabelcounter}

\makeatletter
\newcommand{\labelText}[2]{%
	\refstepcounter{mylabelcounter}%
	\immediate\write\@auxout{%
		\string\newlabel{#2}{{\unexpanded{#1}}{\thepage}{{\unexpanded{#1}}}{mylabelcounter.\number\value{mylabelcounter}}{}}%
	}%
}
\makeatother

\begin{document}
	
	\begin{frontmatter}
		\lstset{basicstyle=\normalsize\ttfamily,breaklines=true}
		
		\title{Explainable e-sports win prediction through Machine Learning classification in streaming\footnote{CC BY-NC-ND license. https://doi.org/10.1016/j.entcom.2025.101027}}
		
		\author[1]{Silvia García-Méndez\corref{mycorrespondingauthor}}\ead{sgarcia@gti.uvigo.es}
		
		\author[1]{Francisco de Arriba-Pérez}\ead{farriba@gti.uvigo.es}
		
		\address[1]{Information Technologies Group, atlanTTic, University of Vigo, Vigo, Spain}
		
		\cortext[mycorrespondingauthor]{Corresponding author: sgarcia@gti.uvigo.es}
		
		\begin{abstract}
			The increasing number of spectators and players in e-sports, along with the development of optimized communication solutions and cloud computing technology, has motivated the constant growth of the online game industry. Even though Artificial Intelligence-based solutions for e-sports analytics are traditionally defined as extracting meaningful patterns from related data and visualizing them to enhance decision-making, most of the effort in professional winning prediction has been focused on the classification aspect from a batch perspective, also leaving aside the visualization techniques. Consequently, this work contributes to an explainable win prediction classification solution in streaming in which input data is controlled over several sliding windows to reflect relevant game changes. Experimental results attained an accuracy higher than \SI{90}{\percent}, surpassing the performance of competing solutions in the literature. Ultimately, our system can be leveraged by ranking and recommender systems for informed decision-making, thanks to the explainability module, which fosters trust in the outcome predictions.
		\end{abstract}
		
		
	\end{frontmatter}
	
	
	\section{Introduction}
	
	Electronic sports (\textit{i.e.}, e-sports) are one of the world's most international and popular sportive events, involving millions of users \citep{Pozzan2022}. The increasing number of spectators and players in e-sports, along with the development of optimized communication solutions and cloud computing technology, has motivated the constant growth of the online game industry \citep{Shin2021} and the increasing popularity of Artificial Intelligence- (\textsc{ai}) based solutions for e-sports analytics \citep{Mora-Cantallops2018}. Notably, the commercial relevance of professional winning prediction is remarkable. In this line, note that this is the most watched part of the game commentator show \citep{Tian2022}.
	
	Most e-sports, particularly regarding the multiplayer online battle arena (\textsc{moba}) e-sports (\textit{e.g.}, Counter-Strike: Global Offensive - \textsc{cs: go}, Defense of the Ancients 2 - \textsc{dota} 2, Heroes of Newerth - \textsc{h}o\textsc{n}, Honor of Kings - \textsc{h}o\textsc{k}, League of Legends - \textsc{l}o\textsc{l}, Tactical Troops: Anthracite Shift - \textsc{tt: as}) attract millions of players and thus, produce a publicly available stream of data regarding matches and competitions. Researchers can easily store and retrieve this vast amount of data to perform e-sports analytics, which can provide valuable tactical knowledge to players \citep{Shin2021}. Regarding \textsc{ai}-based e-sports systems, there exist solutions to advise players on attack tactics and provide recommendations to win the game \citep{Ait-Bennacer2022}, among other valuable applications like tournament classification exploiting Machine Learning (\textsc{ml}) models \citep{Dikananda2021}.
	
	Explainable \textsc{ai} (\textsc{xai}) was created with the ultimate objective of providing transparency to the above \textsc{ml} models that operate as black boxes (\textit{i.e.}, those whose working is unknown by the end users) \citep{akbar2023trustworthy}. Regarding \textsc{xai}, it can be performed following different approaches: (\textit{i}) counterfactual explanations \citep{Wachter2017}, (\textit{ii}) feature importance \citep{Bhatt2020,Vu2021}, (\textit{iii}) natural language descriptions \citep{Ehsan2019}, and (\textit{iv}) visualization \citep{Spinner2019}. Compared to interpretability, which focuses on self-explanatory techniques (\textit{e.g.}, intrinsic interpretable tree-based models), explainability refers to the use of post-hoc techniques, mainly model-agnostic \cite{liu2024leveraging}. Thus, explainability (\textit{e.g.}, local linear and random models like \textsc{lime} and \textsc{shap} \cite{salih2024perspective}) is mostly directed to black-box classification algorithms. Ultimately, the \textsc{ml} models must provide good results and comprehensive interpretation \citep{Boidot2023}. However, even though e-sports analytics is traditionally defined as extracting meaningful patterns from related data and visualizing them to enhance decision-making, most of the effort has been focused on the classification part, overlooking the visualization techniques \citep{Tian2022,Yang2022}. Moreover, the latter classification task is also unexplored on a stream basis.
	
	Consequently, the lack of interpretability and explainability is one of the most significant limitations of previous works focusing on pre-match prediction, which cannot utilize live data streams as input data for the models \citep{hodge2019}. Ultimately, we apply explainability techniques to promote accountability, fairness, interoperability, reliability, and trustworthiness of our solution \cite{Tesfagergish2024}. By these means, we also increase acceptance and adoption of \textsc{ai} by stakeholders, policymakers, online users, and society. Another relevant limitation is that these solutions utilize data from a vast temporal space, which contrasts with the dynamic nature of e-sports. Compared to traditional sports, the core mechanics of e-sports can be significantly altered over time due to continuous updates, making past data obsolete.
	
	In light of the above, the ultimate objective of this work is to create an \textsc{ml}-based win prediction solution for e-sports. Given the characteristics of the use case, the system must operate in real time, and explanations of its operation must be provided to ensure its accountability and reliability and promote its acceptance among end-users. Accordingly, input data is controlled over several sliding windows to reflect relevant game changes. In fact, dynamic data management using sliding windows has been previously applied to different use cases in e-sports apart from the \textsc{moba} genre, which endorses its adaptability and appropriateness \citep{Liaw2020,Lu2022,nicholson2024role}. 
	
	Notably, our work focuses on the tactical first-person shooter video game \textsc{cs: go}, for which players must exhibit good coordination with their team members, low reaction times, and an effective economic strategy to win \citep{Brewer2022}. It represents an appropriate game for e-sports analytics due to its complex round-based mechanics (\textit{e.g.}, map biases, the existence of ties compared to binary-outcome e-sports) that allow studying the interplay of short- and long-term decisions during the game. More specifically, two teams play on either the \textit{counter-terrorist} (\textsc{ct}) or \textit{terrorist} (\textsc{t}) side. Each team comprises five players, and each event consists of 30 rounds with a time limit, starting with a predetermined amount of money. After round 15, the economies are reset since the teams switch sides to reduce the imbalance caused by the map choice. Note that in the start round (\textit{i.e.}, the \textit{pistol round}), the money given to the players is limited, and the latter grows when killing enemies, defusing, or planting a bomb. The team that wins 16 rounds wins the game. In the event of a tie, and if the competition allows it, overtime exists, typically consisting of 6 additional rounds, with map switches occurring every 3 rounds. Overtime ends when a team wins 4 out of 6 rounds in the overtime stage. To increase the value of our proposal, the system was designed to minimize the training data required for win prediction (approximately half the game). 
	
	To summarize, the primary motivation of our work is its application to ranking and recommender systems, which necessitates the existence of an explainability module to promote trust in the outcome predictions. The main objectives of this study are:
	
	\begin{itemize}
		\item Application of streaming techniques for win prediction in e-sports, avoiding the popular, costly batch processing strategy.
		
		\item Generation of a wide range of features to model the evolving nature of e-sport users' behavior and patterns, including analyzing historical data.
		
		\item Creation of natural language descriptions about the predictive behavior of the models included in a graphical dashboard to assist players in real time.
		
		\item The possibility of an expert-in-the-loop to verify and assess the explainability outputs to promote the system's accountability.
		
	\end{itemize}
	
	The rest of this paper is organized as follows. Section \ref{sec:related_work} overviews the relevant competing win prediction systems in the literature regarding the application domain. Section \ref{sec:system} introduces the proposed solution, while Section \ref{sec:results} describes the experimental dataset, implementations, and set-up, along with the empirical evaluation results. Finally, Section \ref{sec:conclusions} concludes and highlights the achievements and future work.
	
	\section{Related work}
	\label{sec:related_work}
	
	In electronic entertainment, e-sports have become viral, thanks to the broadcasting of professional competitions whose audience is constantly rising \citep{Hitar-Garcia2022}. Accordingly, \textsc{cs: go} has recently broken its record player count, reaching \num{1802853} concurrent players on May 2023\footnote{Available at \url{https://www.dexerto.com/csgo/how-many-people-play-csgo-player-count-record-2071859}, June 2025.}. Consequently, e-sports is relevant to the industry and the research community due to its significant commercial size and data value \citep{hodge2019,Reitman2020}. 
	
	Among the most popular applications of e-sports analytics, win prediction, ranking of players \citep{Liu2020}, and prediction of players' skills \citep{Khromov2019} deserve special mention. Win prediction in e-sports is a non-straightforward task that can be addressed from different perspectives \citep{Bahrololloomi2022}. Some authors address it as a binary classification problem (\textit{i.e.}, lose or win); others, like a regression problem in which a constant value is predicted based on the input features \citep{Costa2021}. Regarding input data, existing solutions exploit a wide range of features (\textit{e.g.}, historical match data, player data, and team features) \citep{Bisberg2022}. Primarily, it involves skill modeling, which is relevant to e-sport matchmaking and understanding player collaboration. In this line, note its direct application in ranking players and teams in professional competitions based on historical data for game commentator use cases \citep{bisberg2019}.
	
	In light of the above, \textsc{ml} is an appropriate approach for e-sports analytics due to the availability of high-dimensional (pre-game, in-game, and post-game features related to the match, players, and teams) and high-volume data from \textsc{moba} e-sports \citep{Birant2022}. However, there is still a need to analyze the input features and their relevance towards enhanced win prediction \citep{Costa2021,Ani2019}. Moreover, previous win prediction solutions do not investigate how to interpret the classification outcomes \citep{Yang2022}. Consequently, the lack of interpretability dramatically limits the application of e-sports win prediction solutions in live streaming. Among the features used in e-sports research, \textit{i.e.}, pre-game, in-game, and post-game features, the in-game ones stand out \citep{hodge2019}.
	
	\begin{description}
		\item \textbf{Pre-game features} are those extracted from data before the match starts.
		\item \textbf{In-game features} reflect how the game evolves and are usually exploited for real-time prediction.
		\item \textbf{Post-game features} summarize the match, often at the end of the game, indicating the damages, rewards, and match duration, among other relevant features.
	\end{description}
	
	Table \ref{tab:comparison} summarizes the \textsc{ml}-based \textsc{moba} win prediction solutions from the literature. Note that all works surveyed follow the supervised approach, except for the semi-supervised solution by \citet{Bisberg2022}. Only the system by \citet{hodge2019} applied real-time e-sports analytics. Furthermore, in-game features are broadly used to train the \textsc{ml} models. Finally, note that although few works provide explainability, the authors exclusively explore the feature importance approach \citep{Tian2022,Boidot2023,Yang2022,Hitar-Garcia2022,Bahrololloomi2022,Birant2022,Bahrololloomi2023}.
	
	\begin{table}[!htbp]
		\centering
		\caption{Comparison of \textsc{ml}-based \textsc{moba} win prediction solutions taking into account the e-sports domain, data processing (offline, online), features involved (pre-, in-, post-game), and explainability capability.}
		\label{tab:comparison}
		\begin{tabular}{lcccc}
			\toprule
			\textbf{Authorship} & \textbf{Domain} & \textbf{Process} & \textbf{Features} & \textbf{Explainability}\\
			\midrule
			
			\multirow{2}{*}{\citet{Makarov2018}} & \textsc{dota} 2 & \multirow{2}{*}{Offline} & \multirow{2}{*}{In-game} & \multirow{2}{*}{\xmark} \\
			& \textsc{cs: go} \\
			
			\citet{hodge2019} & \textsc{dota} 2 & Online & In-game & \xmark \\
			
			\citet{Kim2020} & \textsc{l}o\textsc{l} & Offline & In-game & \xmark \\
			
			\citet{Xiao2021} & \textsc{tt: as} & Offline & In-game & \xmark \\
			
			\citet{Bahrololloomi2022} & \textsc{l}o\textsc{l} & Offline & In-game & Feature importance\\
			
			\multirow{2}{*}{\citet{Birant2022}} & \multirow{2}{*}{\textsc{l}o\textsc{l}} & \multirow{2}{*}{Offline} & In-game & \multirow{2}{*}{Feature importance}\\
			& & & Post-game \\
			
			\citet{Bisberg2022} & \textsc{l}o\textsc{l} & Offline & In-game & \xmark \\
			
			\citet{Hitar-Garcia2022} & \textsc{l}o\textsc{l} & Offline & Pre-game & Feature importance \\
			
			\citet{Shen2022} & \textsc{l}o\textsc{l} & Offline & In-game & \xmark \\
			
			\citet{Stanlly2022} & \textsc{dota} 2 & Offline & Pre-game & \xmark \\
			
			\citet{Tian2022} & \textsc{h}o\textsc{k} & Offline & In-game & Feature importance \\
			
			\citet{uddin2022} & \textsc{dota} 2 & Offline & In-game & \xmark \\
			
			\citet{Yang2022} & \textsc{h}o\textsc{k} & Offline & In-game & Feature importance\\
			
			\citet{Bahrololloomi2023} & \textsc{l}o\textsc{l} & Offline & In-game & Feature importance \\
			
			\citet{Boidot2023} & \textsc{l}o\textsc{l} & Offline & In-game & Feature importance \\\midrule
			
			\multirow{3}{*}{\textbf{Proposed}} & \multirow{3}{*}{\textsc{cs: go}} & \multirow{3}{*}{Online} & Pre-game & Feature importance and path\\
			& & & In-game & Natural language description\\
			& & & Post-game & Visual dashboard \\\bottomrule
		\end{tabular}
	\end{table}
	
	As in our case, \citet{Makarov2018} conducted experiments with a \textsc{cs: go} dataset. Additional experiments were performed with \textsc{dota 2} data. Moreover, the authors compared their decision tree-based win prediction model with TrueSkill, a Bayesian rating system\footnote{Available at \url{https://trueskill.org}, June 2025.}. Similar proposals for the \textsc{dota 2} game are those by \citet{Stanlly2022,uddin2022}. Consideration should be given to the work by \citet{Kim2020}, who developed a novel confidence calibration method that takes into account data uncertainty. Conversely, \citet{Xiao2021} designed \textsc{wp-gbdt}, which exploits an ensemble approach combining averaging and stacking for win prediction with gradient boosting.
	
	Furthermore, \citet{Birant2022} proposed a Multi-Objective Multi-Instance Learning (\textsc{momil}) for win prediction. The authors explored multi-instance learning (\textsc{mil}) with five feature vectors for each of the five players in the team, and if they win the game, the sample is associated with the target value 1; otherwise, it is associated with the target value 0. The authors used entropy to measure the impurity of the season-based training data. In contrast, \citet{Bisberg2022} created \textsc{gcn-wp}, an e-sport semi-supervised win prediction solution based on neighborhood classification with convolutional neural networks. Similar proposals are those by \citet{Hitar-Garcia2022,Shen2022}.
	
	In favor of interpretability, \citet{Birant2022} employed the OneR algorithm to assess features' importance according to the minimum error rate on the experimental data. Comparably, feature importance computation by \citet{Hitar-Garcia2022} is based on a decision tree model. Furthermore, \citet{Tian2022} proposed a Hero Featured Network (\textsc{hfn}) that exploits player information. Since the model used is opaque, the Long Short Term Memory network (\textsc{lstm}), the authors decouple the contributions of the input features. In this line, \citet{Yang2022} proposed an interpretable win prediction system exploiting a two-stage spatial-temporal network (\textsc{tsstn}) to extract the contribution of the input features to the final prediction. More recently, \citet{Boidot2023} evaluated explanations generated with \textsc{shap} (a post hoc feature importance explanation method derived from game theory \citep{Lundberg2017}) on an \textsc{l}o\textsc{l} win prediction use case, similar to the earlier proposals by \citet{Bahrololloomi2022,Bahrololloomi2023}.
	
	No related work has applied data management with sliding windows. Outside the \textsc{moba} genre, representative examples are the solutions by \citet{Spijkerman2020} and \citet{Hojaji2024} for the specific case of Formula One and simulated racing, respectively. Like our work, \citet{Scotti2019} utilized data management with sliding windows, in addition to feature engineering, for stream decoding in football matches.
	
	In light of the above, our work addresses the research gap of jointly considering online analytics and providing explainability in win prediction.
	
	\section{System architecture}
	\label{sec:system}
	
\begin{figure}
		\centering
		\includegraphics[scale=0.12]{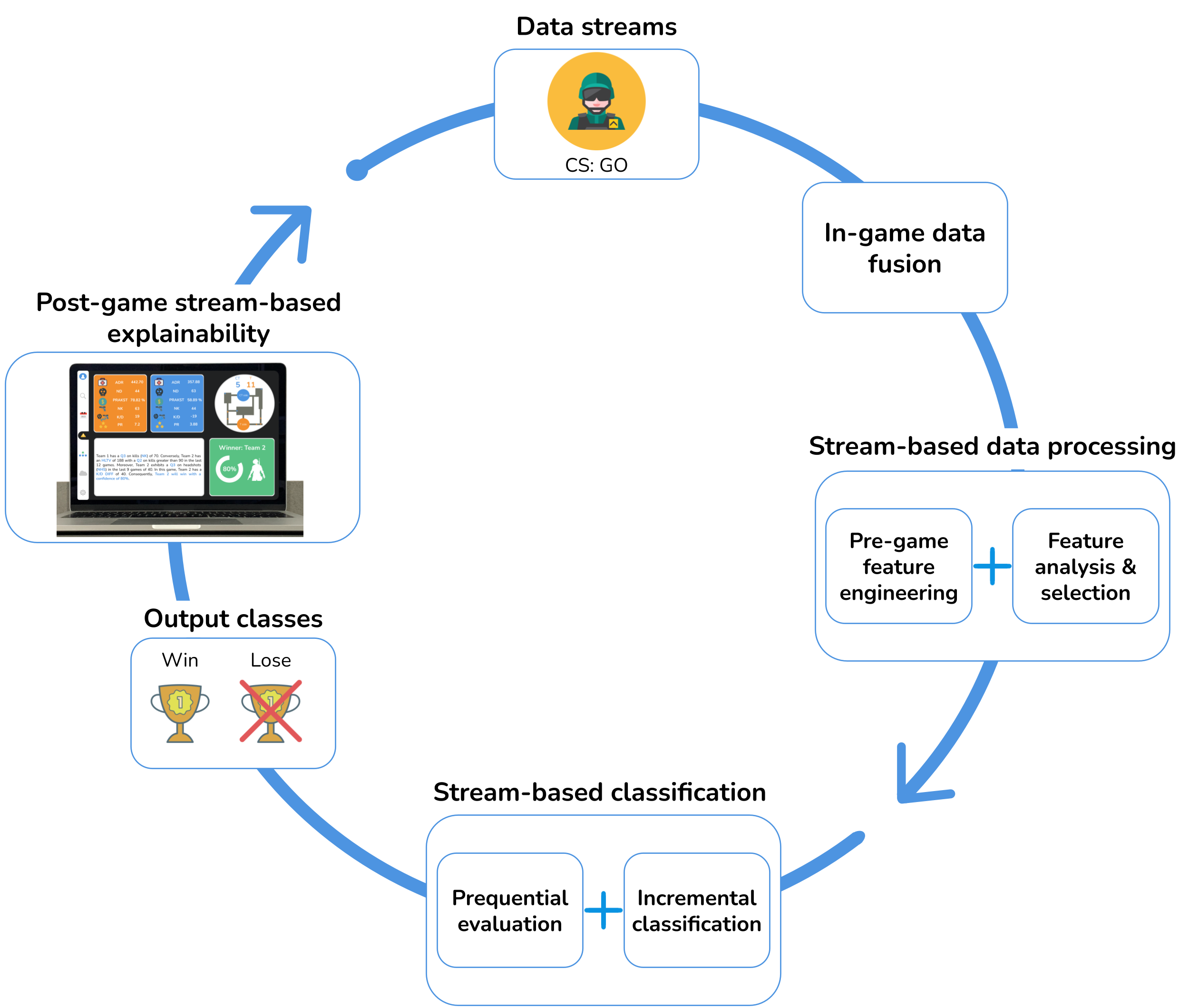}
		\caption{System diagram composed of: (\textit{i}) in-game data fusion, (\textit{ii}) stream-based data processing, (\textit{iii}) stream-based classification, and (\textit{iv}) post-game stream-based explainability.}
		\label{fig:scheme}
	\end{figure}

	Figure \ref{fig:scheme} shows the scheme of the solution. It comprises in-game data fusion (Section \ref{sec:data_fusion}), stream-based data processing (Section \ref{sec:data_processing}), stream-based classification (Section \ref{sec:online_classification}), and post-game stream-based explainability (Section \ref{sec:explainability}).
	
	\subsection{In-game data fusion} 
	\label{sec:data_fusion}
	
	For in-game data fusion, the event identifier, map played, and timestamp will be used to identify a game uniquely and the 5 players per team that participated. The data from each of the 5 players in each team will be aggregated. The latter is especially relevant in e-sports like \textsc{cs: go}, where each player depending on their spam (\textit{i.e.}, where the player appears at the start of the round), weapons, cumulative economy, and abilities, must assume a different role (\textit{entry frager}, \textit{support}, \textit{lurker}, etc.) and therefore their skills will be affected. A snapshot of the player's skills at the team level is computed thanks to the fusion process.
	
	\subsection{Stream-based data processing}
	\label{sec:data_processing}
	
	Data processing is essential to ensure the effective and efficient performance of the \textsc{ml} models in classification problems, even more so in the case of streaming operation. Thus, the system integrates (\textit{i}) pre-game feature engineering to enhance performance and (\textit{ii}) feature analysis and selection to promote efficiency in the proposed solution.
	
	\subsubsection{Pre-game feature engineering}
	\label{sec:feature_engineering}
	
	Players have an individual evolutionary profile that depends on their seniority level (\textit{i.e.}, the number of games played). For this reason, the system builds upon accumulated knowledge over past samples. These samples will be added and removed dynamically, employing sliding windows to compute the distribution function of the number of games per player. The latter computation requires: (\textit{i}) an initial chunk of the streaming data to avoid cold start, (\textit{ii}) the calculation of the number of games per player for that subset, (\textit{iii}) the values of average, 25th, 50th and 75th percentiles (equivalent to the first quartile or \textsc{q1}, \textsc{q2}, and \textsc{q3}) of the game distribution function.
	
\begin{figure}[!htbp]
		\centering
		\includegraphics[scale=0.13]{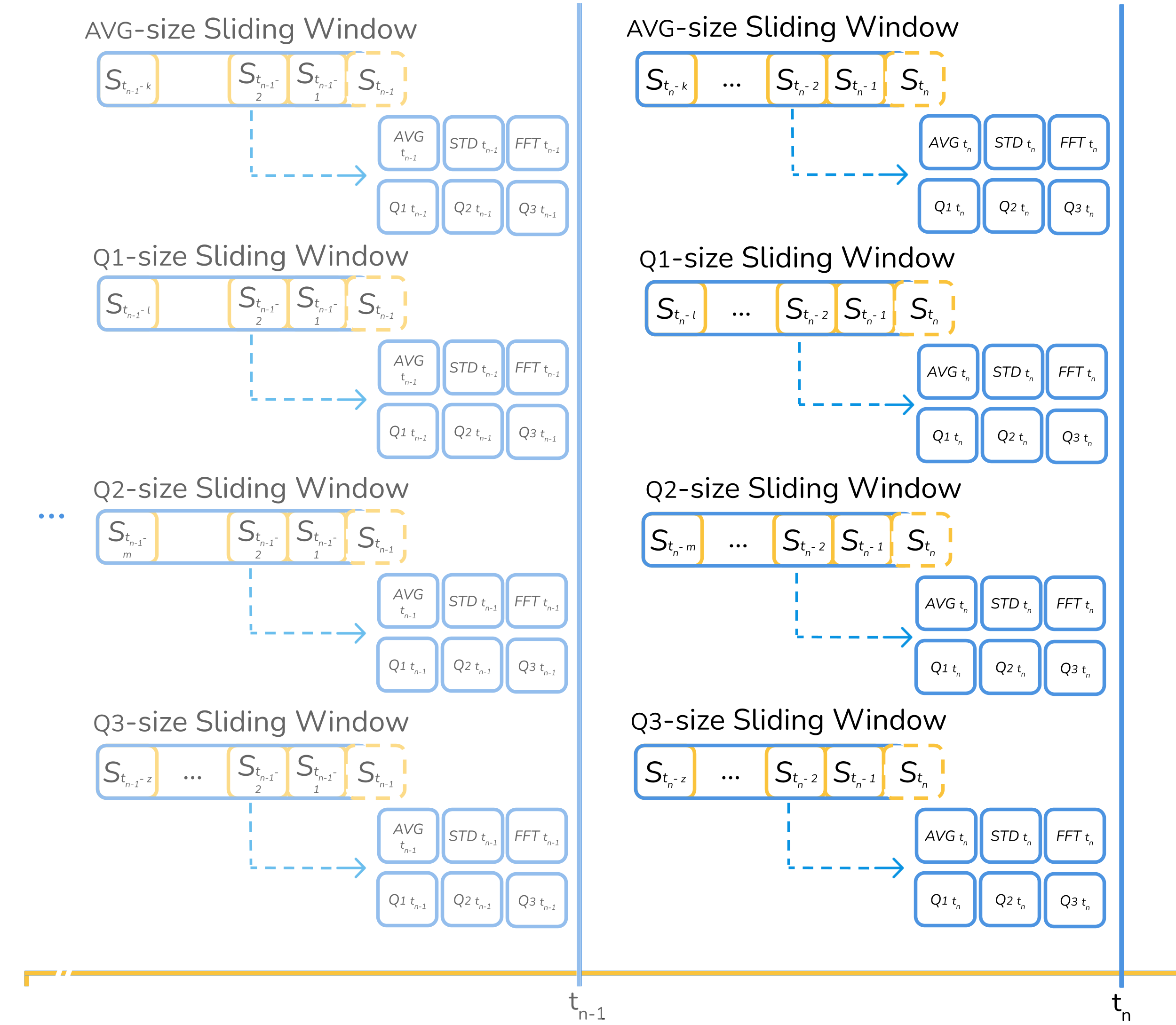}
		\caption{\label{fig:windows}Features engineered using the sliding windows.}
	\end{figure}

	More in detail, four sliding windows will be applied, each with the samples equivalent to the values mentioned above (\textit{i.e.}, average, 25th, 50th, and 75th percentiles). For each of the four sliding windows and each feature in the experimental dataset, some values will be calculated: average, 25th, 50th, and 75th percentiles, the standard deviation (\textsc{std}), and the maximum modulus of the Fast Fourier Transform (\textsc{fft}). Resulting in 6 values per feature and window. It is important to note that not all players will have as much past information as the number of inputs needed for each window. In this case, the window size will be calculated as the minimum between the window size itself and the number of features. Figure \ref{fig:windows}, Algorithm \ref{alg:window_gen}, and Algorithm \ref{alg:feature_eng} illustrate the feature engineering process using sliding windows.
	
	\begin{algorithm*}[!htbp]
		\scriptsize
		\caption{\label{alg:window_gen} {\bf Windows size adjustment.}}
		\DontPrintSemicolon
		\KwIn{Players history.} 
		\KwData{$player\_game\_history\_matrix$}
		
		number\_games\_vector=[ ]
		
		\For {player\_id in set(player\_game\_history\_matrix[``Player ID"])}
		{
			player\_data = player\_game\_history\_matrix[``Player ID"].filter(player\_id)
			
			number\_games\_vector.expand(len(player\_data))  \tcp*[h]{This calculates the number of times a player has played a game and adds it to the vector.}\
		}
		average\_window = number\_games\_vector.mean()
		
		q1\_window = number\_games\_vector.percentil(25)
		
		q2\_window = number\_games\_vector.percentil(50)
		
		q3\_window = number\_games\_vector.percentil(75)
		
		\KwRet{$[average\_window, q1\_window, q2\_window, q3\_window]$} \tcp*[h]{Returns four windows sizes for feature engineering.}\;
		
	\end{algorithm*}
	
	\begin{algorithm*}[!htbp]
		\scriptsize
		\caption{\label{alg:feature_eng} {\bf Feature engineering.}}
		\DontPrintSemicolon
		\KwIn{Player ID, player history, and windows sizes.} \tcp*[h]{This method is executed for each player involved in a game.}\
		
		\KwData{$player\_id, player\_game\_history\_matrix$, $average\_window$, $q1\_window$, $q2\_window$, $q3\_window$}
		
		player\_data = player\_game\_history\_matrix[``Player ID"].filter(player\_id)
		
		statistic\_features\_vector=[ ]
		
		\For {window in [average\_window, q1\_window, q2\_window, q3\_window]}
		{
			\uIf{$window > len(player\_data$)} {
				window = len(player\_data)
			}
			
			\For {column in player\_data.columns}
			{
				column\_values\_vector = player\_data[column][0:window]
				
				$[average, maximum, minimum, q1, q2, q3, std, fft]$ = feature\_statistics\_eq1(column\_values\_vector) \tcp*[h]{The method of calculation is described in Equation 1.}\
				
				statistic\_features\_vector.expand([$mean, maximum, minimum, q1, q2, q3, std, fft$])
			}
		}
		\KwRet{$statistic\_features\_vector$} \tcp*[h]{Returns all the features generated for each statistic and sliding window.}\
		
	\end{algorithm*}
	
	Equation \eqref{eq:feature_engineering} shows how the statistic values for the features are computed, where $n$ is the number of games of a player and $X[n]$ is the historical feature in the last $n$ player games. Consequently, $Y[n]$ is the ordered version of $X[n]$.
	
	\begin{equation}\label{eq:feature_engineering}
		\begin{split}
			\forall n \in \{1 ... \infty\}\\ \\
			X[n] = \{ x[0],\ldots,x[n]\}. \\
			Y[n] = \{y_0[n], y_1[n],\ldots,y_{n-1}[n]\} \mid  y_0[n]\leq y_1[n]\leq\ldots\leq y_{n-1}[n], \\
			\mbox{where} ; \forall x \in X[n], \; x \in Y[n]. \\ \\
			avg[n]=\frac{1}{n}\sum_{i=0}^{n} y_i [n]\\
			Q_{1}[n]=y_{\lfloor\frac{1}{4}n\rceil}[n] \\
			Q_{2}[n]=y_{\lfloor\frac{2}{4}n\rceil} [n]\\
			Q_{3}[n]=y_{\lfloor\frac{3}{4}n\rceil} [n]\\
			std[n]=\sigma(X[n]) \\
			F[n]=|FFT(X[n])|_\infty\\
		\end{split}
	\end{equation}
	
	\subsubsection{Feature analysis \& selection}
	\label{sec:feature_analysis_selection}
	
	Reducing the number of features used for training the \textsc{ml} models is vital in every classification problem \citep{Shaddeli2023}. The latter becomes essential in the stream-based scenario to ensure the effective and efficient performance of the proposed solution \citep{Gomes2019}. Moreover, the selected features may vary since the system operates in a streaming environment. 
	
	Low-relevant features will be removed, exploiting the variance thresholding approach for feature selection based on variance analysis. Thus, features with low variance are discarded, inspired by the literature, which considers these features to be lowly relevant for classification purposes \citep{Treistman2022}. Firstly, the variance of the features in each incoming sample is computed. Only the features that meet a configurable threshold are considered in the following stages. More in detail, the variance threshold is obtained using the cold start subset mentioned in Section \ref{sec:feature_engineering} as required in the streaming scenario. Then, the variance of the features in that subset is calculated. Finally, the value of the 10th percentile is selected as the threshold, indicating the value from which the feature should be considered an outlier.
	
	\subsection{Stream-based classification}
	\label{sec:online_classification}
	
\begin{figure*}[!htbp]
		\centering
		\includegraphics[scale=0.15]{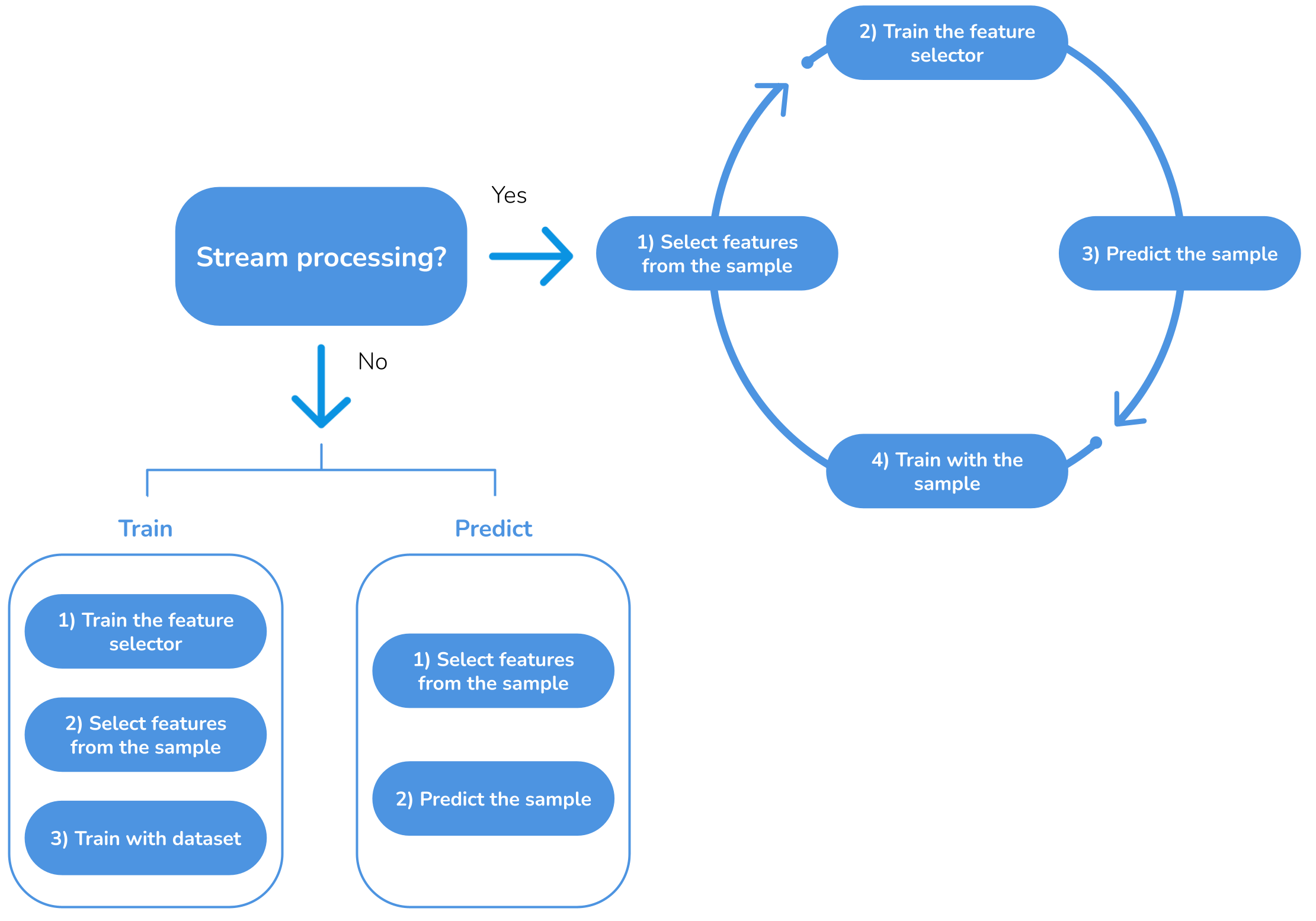}
		\caption{\label{fig:streaming_batch} Streaming and batch supervised classification.}
	\end{figure*}

	Figure \ref{fig:streaming_batch} shows the supervised classification pipeline in streaming and batch. In contrast to batch learning, the first step involves feature selection, and the selector update occurs with every incoming sample, ordered by their timestamp. Ultimately, the sample is classified, and the model is updated according to this current sample. The main difference between the two approaches is that the prediction in batch occurs after selector update and feature extraction. Conversely, the prediction comes first in stream-based classification, and then the model is re-trained. Additionally, in the batch scheme, the training process involves the entire dataset, which prevents the model from undergoing a scalable, continuous update.
	
	Consequently, in our study, the \textsc{ml} models are first tested using prequential evaluation and then trained on the incoming samples. Note that data decimation is applied in training and testing; one in ten samples is selected, and the rest are discarded. 
	
	Below, the \textsc{ml} models used for classification are listed. Note that these models were selected inspired by the revision of the competing works in the literature \citep{Boidot2023,hodge2019,Bahrololloomi2023,Makarov2018}. Moreover, deep learning models were excluded due to the limitations related to fixed network weights after training, which prevents their application in a streaming scenario \cite{Protonotarios2022} and its black-box nature since they are not intrinsically interpretable like tree-based models \cite{Qamar2023}.
	
	\begin{itemize}
		
		\item \textbf{Gaussian Naive Bayes} (\textsc{gnb}) \citep{Tieppo2021}. It is based on the traditional Gaussian distribution from the Naive Bayes (\textsc{nb}) model, which is appropriate for stream-based classification. It is used as a baseline. 
		
		\item \textbf{Hoeffding Adaptive Tree Classifier} (\textsc{hatc}) \citep{Mrabet2019}. This model is based on an online single tree with a mechanism to evaluate branch performance.
		
		\item \textbf{Adaptive Random Forest Classifier} (\textsc{arfc}) \citep{Fatlawi2020}. It is similar to the latter model but employs an ensemble strategy that comprises resampling and random feature selection. Prediction is obtained by computing majority voting.
		
	\end{itemize}
	
	\subsection{Post-game stream-based explainability}
	\label{sec:explainability}
	
	For end-users in a real-time gaming context, explainability can help them be accountable for their performance, anticipate potential rewards, or be prepared for penalties in sporting events. The same applies to identifying areas for improvement in the game for future events and detecting risky patterns and behaviors that may lead to losing the match or tournament. At the same time, the explanations provided may help e-sport users optimize the resources available (\textit{e.g.}, team selection and economic strategy). In short, \textsc{xai} can promote player reflection, learning, and immersion in the game.
	
	Accordingly, post-game features are exploited as part of the information provided to the end users in the explainability dashboard. To gather the relevant data to be included in the natural language descriptions, the decision path (\textit{greater than} and \textit{lower than} branches) is traversed to extract the track and the feature occurrence number (\textit{i.e.}, frequency of the features). A single or configurable number of estimators is considered for the \textsc{hatc} and the \textsc{arfc} models, respectively. The latter is used to order the features to select the most important for classification purposes. Moreover, the performance of the \textsc{ml} model is also included as part of the natural language description and explainability dashboard using the statistical parameters regarding the sliding windows (average, \textsc{q\textsubscript{1}}, \textsc{q\textsubscript{2}}, \textsc{q\textsubscript{3}}, \textsc{std} and \textsc{fft} signal). Algorithm \ref{alg:explainability} summarizes the process.
	
	\begin{algorithm*}[ht!]
		\scriptsize
		\caption{Explainability process.}\label{alg:explainability}
		\DontPrintSemicolon
		\KwIn{Estimator and sample.} \tcp*[h]{This process is repeated for each estimator in the tree classifier.}\
		
		\KwData{$estimator$, $sample$}
		\Begin{
			
			$feature\_frequency\_vector=[]$
			
			$tree\_node=estimator[0]$; \tcp*[h]{Root node.}\
			
			\While{$tree\_node \neq nil$}
			{
				$feature= tree\_node[feature]$
				
				$threshold = tree\_node[threshold]$
				
				$feature\_frequency\_vector.append(feature)$
				
				\uIf{$sample[feature] \leq threshold$} {
					
					$tree\_node=tree\_node[left\_branch]$
				}
				\uElse {
					
					$tree\_node=tree\_node[right\_branch]$
				}
				
			}
		}
		
		feature\_frequency\_vector = feature\_frequency\_vector.counter().values()
		
		\KwRet{$feature\_frequency\_vector[0:5]$} \tcp*[h]{Returns first 5 features sorted by frequency of occurrence.}\
	\end{algorithm*}
	
	\section{Evaluation and discussion}
	\label{sec:results}
	
	All the experiments were executed on a computer with the following specifications:
	\begin{itemize}
		\item \textbf{Operating System}: Ubuntu 18.04.2 LTS 64 bits
		\item \textbf{Processor}: Intel\@Core i9-10900K \SI{2.80}{\giga\hertz}
		\item \textbf{RAM}: \SI{96}{\giga\byte} DDR4 
		\item \textbf{Disk}: \SI{480}{\giga\byte} NVME + \SI{500}{\giga\byte} SSD
	\end{itemize}
	
	\subsection{Experimental dataset}
	\label{sec:experimental_data_Set}
	
	The publicly available dataset\footnote{\label{data_link}Available at \url{https://www.kaggle.com/datasets/mateusdmachado/csgo-professional-matches}, June 2025.} comprises information about the players' skills (subset \textsc{p}, \num{383317} samples) and the results of the \textsc{cs: go} (subset \textsc{r}, \num{45773} samples), in separate \texttt{csv} files. For clarity, note that an event comprises up to 3 games (in different maps) lasting between 16 and 30 rounds.
	
	Each sample in the subset \textsc{p} constrains players' metrics regarding each event (see Table \ref{tab:features_players}, raw features). Conversely, each sample in the subset \textsc{r} represents a game between two teams (see Table \ref{tab:features_results}, raw features). Summarizing, the experimental data are organized at the team (team 1 and team 2) and side (\textsc{ct} and \textsc{t}) levels. Using the feature \#7 in Table \ref{tab:features_results}, the data in both subsets were assigned to a specific team and part of the game (first or second half). If team 1 or 2 starts in the \textsc{ct} side in the first half of the game, it will play in the \textsc{t} side in the second half. Features \#11 and \#12 in Table \ref{tab:features_players} and feature \#11 in Table \ref{tab:features_results} are engineered with the latter transformation.
	
	Moreover, the players' skills comprise:
	\begin{description}
		\item \textbf{Basic skills}: (\textit{i}) average damage per round (\textsc{adr}), (\textit{ii}) the number of deaths (\textsc{nd}), (\textit{iii}) percent of rounds the player assisted, killed, survived or was traded (\textsc{prakst}), (\textit{iv}) the number of kills (\textsc{nk}), (\textit{v}) difference between deaths and kills (\textsc{k/d diff}), (\textit{vi}) rating in the game (\textsc{pr}).
		
		\item \textbf{Advanced skills}: (\textit{i}) the number of assists (\textsc{nass}), (\textit{ii}) the number of flash assists (\textsc{nfa}), (\textit{iii}) the number of headshots (\textsc{nhs}), (\textit{iv}) difference of the cumulative first to kill and first to die per round (\textsc{fk/fd diff}).
	\end{description}
	
	Samples in both subsets were selected based on their date, resulting in \num{376469} and \num{45079} samples for the subset \textsc{p} and \textsc{r}, respectively. The number of players involved is \num{12153}, and the average of games per player is 3.71. Table \ref{tab:dataset_distribution} summarizes the distribution of the samples per target class in our classification problem. 
	
	\begin{table*}[!htbp]
		\centering
		\footnotesize
		\caption{\label{tab:features_players}Features in subset \textsc{p} by type (raw from the original dataset or engineered - Engi.).}
		\begin{tabular}{ccp{4.5cm}p{4.5cm}}
			\toprule
			\textbf{Num.} & \bf Type & \bf Name & \bf Description\\\midrule
			1 & \multirow{12}{*}{Raw} & Date & Date of the event.\\
			2 & & Event \textsc{id} & \textsc{id} of the event.\\
			3 & & Game \textsc{id} & \textsc{id} of the game.\\
			4 & & Team \textsc{id} & \textsc{id} of the team.\\
			5 & & Player \textsc{id} & \textsc{id} of the player.\\
			6 & & Map\{1,2,3\} & \textsc{id} of the map.\\
			7 & & Skills & Basic \& advanced skills for all maps.\\
			8 & & Skills by side\{\textsc{ct},\textsc{t}\} & Basic skills for all maps by side.\\
			9 & & Skills by map\{1,2,3\} & Basic \& advanced skills in the map.\\
			10 & & Skills by map\{1,2,3\} and side\{\textsc{ct},\textsc{t}\} & Basic skills in the map by side.\\\midrule
			11 & \multirow{4}{*}{Engi.} & Skills by half\{1,2\} & Basic skills for all maps by game half.\\
			12 & & Skills by map\{1,2,3\} and half\{1,2\} & Basic skills in the map and game half.\\
			\bottomrule
		\end{tabular}
	\end{table*}
	
	\begin{table*}[!htbp]
		\centering
		\footnotesize
		\caption{\label{tab:features_results}Features in subset \textsc{r} by type (raw from the original dataset or engineered - Engi.).}
		\begin{threeparttable}
			\begin{tabular}{ccp{4cm}p{5cm}}
				\toprule
				\textbf{Num.}&\textbf{Type} & \bf Name & \bf Description\\\midrule
				1 &\multirow{11}{*}{Raw}& Date & Date of the event.\\
				2 & & Event \textsc{id} & \textsc{id} of the event.\\
				3 & & Game \textsc{id} & \textsc{id} of the game.\\
				4 & & Team \textsc{id} & \textsc{id} of the team.\\
				5 & & Map\{1,2,3\} & \textsc{id} of the map.\\
				6 & & \textsc{hltv} rank by team\{1,2\} & Team Half-Life Television ranking\tnote{1}.\\
				7 & & Starting \textsc{ct} side & Team \textsc{id} in the \textsc{ct} side in the first half.\\
				8 & & Score by team\{1,2\} & Number of rounds won by team.\\
				9 & & Score by team\{1,2\} and side\{\textsc{ct},\textsc{t}\} & Number of rounds won by team and side.\\
				10 & & Target & The \textsc{id} of the team that won the game.\\\midrule
				\multirow{2}{*}{11} & \multirow{2}{*}{Engi.} & Score by team\{1,2\} and half\{1,2\} & Number of rounds won by team and half.\\
				\bottomrule
			\end{tabular}
			\begin{tablenotes}
				\item[1] Available at \url{https://www.hltv.org}, June 2025.
			\end{tablenotes}
		\end{threeparttable}
	\end{table*}
	
	\begin{table}[!htbp]
		\centering
		\caption{\label{tab:dataset_distribution}Distribution of classes in the experimental dataset.}
		\begin{tabular}{lS[table-format=6.0]}
			\toprule \textbf{Winner} & \multicolumn{1}{c}{\textbf{Number of samples}}\\ \midrule
			Team 1 & 24236 \\
			Team 2 & 20843 \\
			\midrule
			Total & 45079 \\ \bottomrule
		\end{tabular}
	\end{table}
	
	\subsection{In-game data fusion} 
	\label{sec:data_fusion_results}
	
	To fusion the in-game features in the \textsc{p} and \textsc{r} subsets, the features \#1, \#2, \#3, and \#6 / \#5 (map \textsc{id}) in Table \ref{tab:features_players} and Table \ref{tab:features_results} are used as the key. The latter selection results in 10 (5 players in each team) and 1 sample of the \textsc{p} and \textsc{r} subsets, respectively.
	
	To extract the skills related to each team, we used the feature \#4 in both Table \ref{tab:features_players} and Table \ref{tab:features_results} as key, and then applied Equation \ref{eq:fusion} using feature \#12 in Table \ref{tab:features_players} and feature \#11 in Table \ref{tab:features_results} (h1 and h2 stand for the first and second half of the game).
	
	\begin{equation}
		\label{eq:fusion}
		skills_{team}= feature \#12 (h1) + \frac{feature \#12 (h2)}{feature \#11 (h2)}
	\end{equation}
	
	\begin{table*}[!htbp]
		\centering
		\footnotesize
		\caption{\label{tab:features_fusion}Features used for training and testing after data fusion and processing.}
		\begin{tabular}{ccp{3cm}p{5.5cm}}
			\toprule
			\textbf{Num.} & \bf Type & \bf Name & \bf Description\\\midrule
			1 & \multirow{7}{*}{In-game} & Date & Date of the event.\\
			2 & & Map\{1,2,3\} & \textsc{id} of the map.\\
			\multirow{2}{*}{3} & & Score by team and side in 1st half & Number of rounds won by team and side in the first half of the game.\\
			4 & & \textsc{hltv} rank by team & Team Half-Life Television ranking.\\
			\multirow{2}{*}{5} & & \multirow{2}{*}{Skills by team} & Aggregation of basic skills in the map by side in the first 16 rounds.\\\midrule 
			\multirow{3}{*}{6} & \multirow{3}{*}{Pre-game} & \multirow{3}{*}{Sliding windows skills} & Sliding windows features of basic and advanced skills for all maps (average, 25th, 50th, 75th, \textsc{std} and \textsc{fft}).\\
			\bottomrule
		\end{tabular}
	\end{table*}
	
	\subsection{Stream-based data processing}
	\label{sec:data_processing_results}
	
	In this section, the implementations used and the results obtained for (\textit{i}) pre-game feature engineering and (\textit{ii}) feature analysis and selection stages are presented and discussed.
	
	\subsubsection{Pre-game feature engineering}
	\label{sec:feature_engineering_results}
	
\begin{figure}[!htbp]
		\centering
		\includegraphics[scale=0.3]{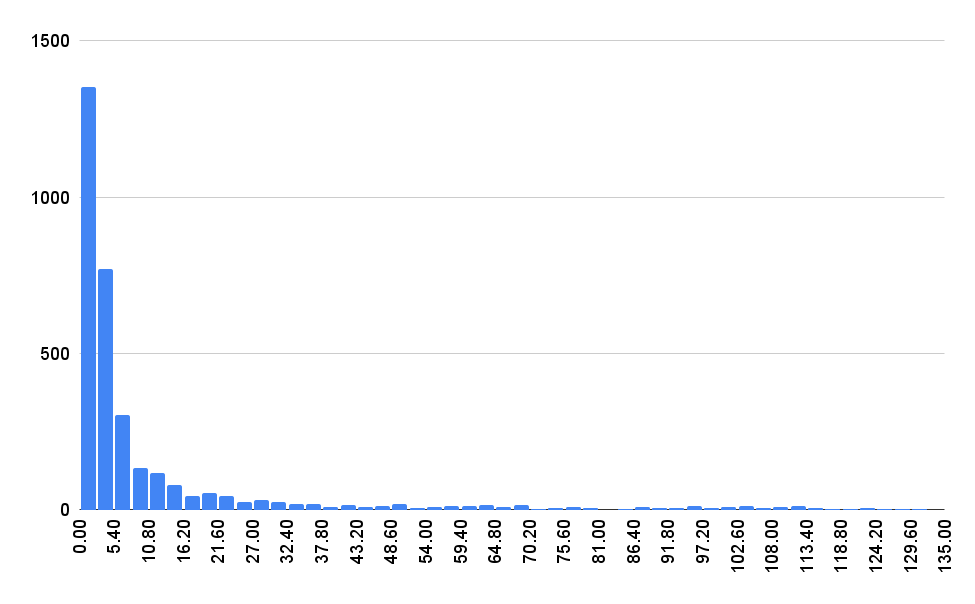}
		\caption{Distribution function of the number of games per player.}
		\label{fig:players_distribution}
	\end{figure}

	The average, 25th, 50th, and 75th percentiles of the cold start data (\num{4507} samples) correspond to 12, 2, 3, and 9. The latter values enable the system to engineer cumulative features for both basic and advanced players' skills (see feature \#7 in Table \ref{tab:features_players}). Figure \ref{fig:players_distribution} shows the distribution function of the number of games per player, which follows an exponentially decreasing tendency.
	
	The in-game variables obtained in the data fusion stage (see Section \ref{sec:data_fusion}) are then merged with the pre-game features engineered (see Section \ref{sec:feature_engineering}), resulting in 279 features (39 in-game and 240 pre-game features). Table \ref{tab:features_fusion} summarizes the in-game and pre-game features used. These will be the input features used to train the stream-based classification models.
	
	\subsubsection{Feature analysis \& selection}
	\label{sec:feature_analysis_selection_results}
	
	To evaluate the cumulative variance of the features engineered for each incoming sample, the solution employs \texttt{VarianceThreshold}\footnote{Available at \url{https://riverml.xyz/0.11.1/api/feature-selection/VarianceThreshold}, June 2025.} from \texttt{River}\footnote{Available at \url{https://riverml.xyz/0.11.1}, June 2025.}. The selection threshold for scenario \textsc{a} and \textsc{b}, explained below, are 0.004 and 0.076, respectively.
	
	\subsection{Stream-based classification}
	\label{sec:online_classification_result}
	
	Two experimental scenarios were designed.
	
	\begin{itemize}
		\item \textbf{Scenario A}. It comprises the features \#1-\#4 from Table \ref{tab:features_fusion} to evaluate the effect of in-game features regarding team data.
		
		\item \textbf{Scenario B}. It comprises all the features from Table \ref{tab:features_fusion} to evaluate the combined effect of pre-game and in-game data, including players' skills.
	\end{itemize}
	
	{\sc ml} models were incrementally trained and evaluated using an \textit{ad hoc} implementation of \texttt{EvaluatePrequential}\footnote{Available at \url{https://scikit-multiflow.readthedocs.io/en/stable/api/generated/skmultiflow.evaluation.EvaluatePrequential.html}, June 2025.}.
	
	The online supervised classification models used are \textsc{gnb}\footnote{Available at \url{https://riverml.xyz/dev/api/naive-bayes/GaussianNB}, June 2025.}, \textsc{hatc}\footnote{Available at \url{https://riverml.xyz/0.11.1/api/tree/HoeffdingAdaptiveTreeClassifier}, June 2025.} and \textsc{arfc}\footnote{Available at \url{https://riverml.xyz/0.11.1/api/ensemble/AdaptiveRandomForestClassifier}, June 2025.} from \texttt{River}. Listings \ref{hatc_conf} and \ref{arfc_conf} show the hyper-parameter optimization ranges used for the \textsc{hatc} and \textsc{arfc} models, respectively. Note that the \textsc{gnb} model has no hyper-parameters to tune. The selected hyper-parameters are:
	
	\begin{description}
		\item \textbf{HATC}: [50, 0.5, 50], [50, 0.5, 50] for scenario \textsc{a} and scenario \textsc{b}, respectively.
		\item \textbf{ARFC}: [100, 50, 10], [50, 50, 10] for scenario \textsc{a} and scenario \textsc{b}, respectively.
	\end{description}
	
	\begin{lstlisting}[frame=single,caption={\textsc{hatc} hyperparameter configuration.},label={hatc_conf},emphstyle=\textbf,escapechar=ä]
		depth = [50, 100, 200]
		tiethreshold = 0.5, 0.05, 0.005]
		maxsize = [50, 100, 200]
	\end{lstlisting}
	
	Table \ref{tab:classification_results} shows the results obtained with the \textsc{ml} models selected in the experimental scenarios mentioned above. For the first scenario, \textsc{a}, the performance difference between models is not statistically significant (\textit{i.e.}, less than \SI{5}{\percent}). The evaluation metrics are promising between \SI{68}{\percent} and \SI{79}{\percent}. Moreover, in scenario \textsc{b}, the effect of combining pre-game and in-game features, including data related to players' skills, improves the results for \textsc{hatc} and \textsc{arfc}. Unfortunately, the simplicity of the baseline \textsc{gnb} model prevents it from taking advantage of the latter features. In this regard, the biggest difference are between the \textsc{gnb} baseline and \textsc{arfc} (\textit{i.e.}, \SI{18.34}{\percent}, \SI{16.98}{\percent}, \SI{17.37}{\percent}, \SI{18.33}{\percent} in accuracy, presion, recall and \textsc{f}-measure, respectively). Since it is a baseline model, \textsc{gbn} is 40 times faster than \textsc{arfc}. In the end, the \textsc{arfc} classifier exhibits the best performance (\textit{i.e.}, all metrics are above \SI{90}{\percent}). Even though it has a high computational cost, this does not affect the stream operation of the win prediction solution (\textit{i.e.}, real-time processing is achieved since the model can process 22.5 samples per second). Note that the latter model is interpretable by traversing the branches of its decision trees.
	
	\begin{lstlisting}[frame=single,caption={\textsc{arfc} hyperparameter configuration.},label={arfc_conf},emphstyle=\textbf,escapechar=ä]
		models = [25, 50, 100]
		features = [50, 150, 300]
		lambda = [10, 50, 100]
	\end{lstlisting}
	
	\begin{table*}[!htbp]
		\centering
		\footnotesize
		\caption{\label{tab:classification_results}Win prediction results in streaming.}
		\begin{tabular}{ccccccccccccS[table-format=3.2]}
			\toprule
			\bf Sce. & \bf Model & \bf Acc. & \multicolumn{3}{c}{\bf Precision} & \multicolumn{3}{c}{\bf Recall} & \multicolumn{3}{c}{\bf F-measure} &{\bf Time (s)}\\
			\cmidrule(lr){4-6}
			\cmidrule(lr){7-9}
			\cmidrule(lr){10-12}
			& & & Macro & \#1 & \#2 & Macro & \#1 & \#2  & Macro & \#1 & \#2 \\
			\midrule
			
			\multirow{3}{*}{A} & 
			\textsc{gnb} & 74.97 & 74.84 & 75.85 & 73.82 & 74.59 & 79.08 & \bf70.10 & 74.67 & 77.43 & 71.91 & 1.15\\
			& 
			\textsc{hatc} & 74.02 & 73.89 & 74.82 & 72.95 & 73.59 & 78.59 & 68.59 & 73.68 & 76.66 & 70.70 &2.21\\
			& 
			\textsc{arfc} & \bf74.99 & \bf74.86 & \bf75.86 & \bf73.86 & \bf74.61 & \bf79.12 & \bf70.10 & \bf74.69 & \bf77.46 & \bf71.93 & 144.72\\
			
			\cmidrule(lr){2-13}
			
			\multirow{3}{*}{B} & 
			\textsc{gnb} & 74.17 & 75.60 & 83.88 & 67.33 & 74.98 & 64.38 & 85.58 & 74.11 & 72.85 & 75.36 & 5.33\\
			& 
			\textsc{hatc} & 84.61 & 84.51 & 85.91 & 83.11 & 84.54 & 85.41 & 83.67 & 84.53 & 85.66 & 83.39 & 13.00\\
			& 
			\textsc{arfc} & \bf92.51 & \bf92.58 & \bf91.87 & \bf93.30 & \bf92.35 & \bf94.44 & \bf90.25 & \bf92.44 & \bf93.14 & \bf91.75 & 200.58\\
			\bottomrule
		\end{tabular}
	\end{table*}
	
	To further evaluate the robustness of the proposed streaming solution, we apply a segment-wise validation strategy following the nested cross-validation method used in continuous learning\footnote{Available at \url{https://scikit-learn.org/stable/auto_examples/model_selection/plot_nested_cross_validation_iris.html}, June 2025.}. For this purpose, the data stream is divided into sequential, non-overlapping blocks of \num{10000} instances. In each block, the classifier and feature selector are restarted, and their performance is evaluated following the prequential scheme (prediction before training). This procedure allows us to analyze the model's performance across the stream. As it can be observed in Table \ref{tab:classification_nested_results}, \textsc{arfc} is the best-performing classifier with values around \SI{90}{\percent} in all evaluation metrics, showing a performance reduction of less than \SI{5}{\percent} compared to traditional streaming evaluation in Table \ref{tab:classification_results}. This additional validation contributes to a more realistic assessment of system behavior when data is not available simultaneously but instead arrives incrementally and when models can only access partial historical data.
	
	\begin{table*}[!htbp]
		\centering
		\footnotesize
		\caption{\label{tab:classification_nested_results}\textcolor{black}{Win prediction results in streaming using nested cross-validation (scenario B).}}
		\begin{tabular}{ccccccccccccS[table-format=3.2]}
			\toprule
			\bf Model & \bf Acc. & \multicolumn{3}{c}{\bf Precision} & \multicolumn{3}{c}{\bf Recall} & \multicolumn{3}{c}{\bf F-measure} &{\bf Time (s)}\\
			\cmidrule(lr){3-5}
			\cmidrule(lr){7-8}
			\cmidrule(lr){9-11}
			& & Macro & \#1 & \#2 & Macro & \#1 & \#2  & Macro & \#1 & \#2 \\
			\midrule
			
			\textsc{gnb} & 74.12 & 74.43 & 79.69 & 69.18 & 74.49 & 69.65 & 79.33 & 74.12 & 74.34 & 73.90 & 4.13\\
			& 
			\textsc{hatc} & 75.88 & 75.79 & 76.57 & 75.00 & 75.58 & 79.49 & 71.67 & 75.65 & 78.00 & 73.30 & 9.88\\
			& 
			\textsc{arfc} & \bf88.87 & \bf88.93 & \bf88.44 & \bf89.42 & \bf88.68 & \bf91.25 & \bf86.10 & \bf88.78 & \bf89.82 & \bf87.73 & 170.33\\
			\bottomrule
		\end{tabular}
	\end{table*}
	
	Finally, Table \ref{tab:comparison_results} summarizes the performance comparison between our solution and the literature. Only \citet{Makarov2018} applied win prediction in \textsc{cs: go}. The authors did not publicly provide their experimental dataset, which prevented evaluation of our solution, and \citet{hodge2019} operated on a stream-based approach. As a result of the theoretical comparison, our solution's accuracy exceeds 30.51 and 15 percent points for the above solutions, respectively. 
	
	\begin{table}[!htbp]
		\centering
		\caption{\label{tab:comparison_results}Performance comparison with the literature.}
		\begin{tabular}{lcccc}
			\toprule
			\bf Authorship & \bf Domain & \bf Processing & \bf Accuracy & Explainability\\
			\midrule
			\citet{Makarov2018} & \textsc{cs: go} & Offline & \SI{62.00}{\percent} & \xmark\\
			\citet{hodge2019} & \textsc{dota}2 & Online & \SI{77.51}{\percent} & \xmark\\\midrule
			\multirow{3}{*}{Proposal} & \multirow{3}{*}{\textsc{cs: go}} & \multirow{3}{*}{Online} & \multirow{3}{*}{\SI{92.51}{\percent}} & Feature importance and path\\
			& & & & Natural language description\\
			& & & & Visual dashboard\\
			\bottomrule
		\end{tabular}
	\end{table}
	
	Regarding the scalability of the proposed solution, it simulates streaming operations by following a single-thread scheme. Parallel computation, where several models simultaneously serve the requests, will considerably reduce its response time. Elastic hardware solutions, like distributed infrastructure, would be appropriate. Moreover, technologies such as Apache Kafka and Spark can also adequately manage the data stream. However, the replica models would not have a complete picture of the incoming data in this context. Alternatively, allocating threads could be based on specific competitions and platforms. More specifically, possible challenges lie in the heterogeneity of e-sports genres regarding performance indicators and game mechanics, which necessitate that the solution be able to integrate multimodal data in real time. About the \textsc{api}s employed for data gathering, consideration should be given to their latency. Ultimately, scaling the solution across multiple games or platforms would benefit from its modular design.
	
	\subsection{Post-game stream-based explainability}
	\label{sec:explainability_results}
	
\begin{figure}
		\centering
		\includegraphics[scale=0.12]{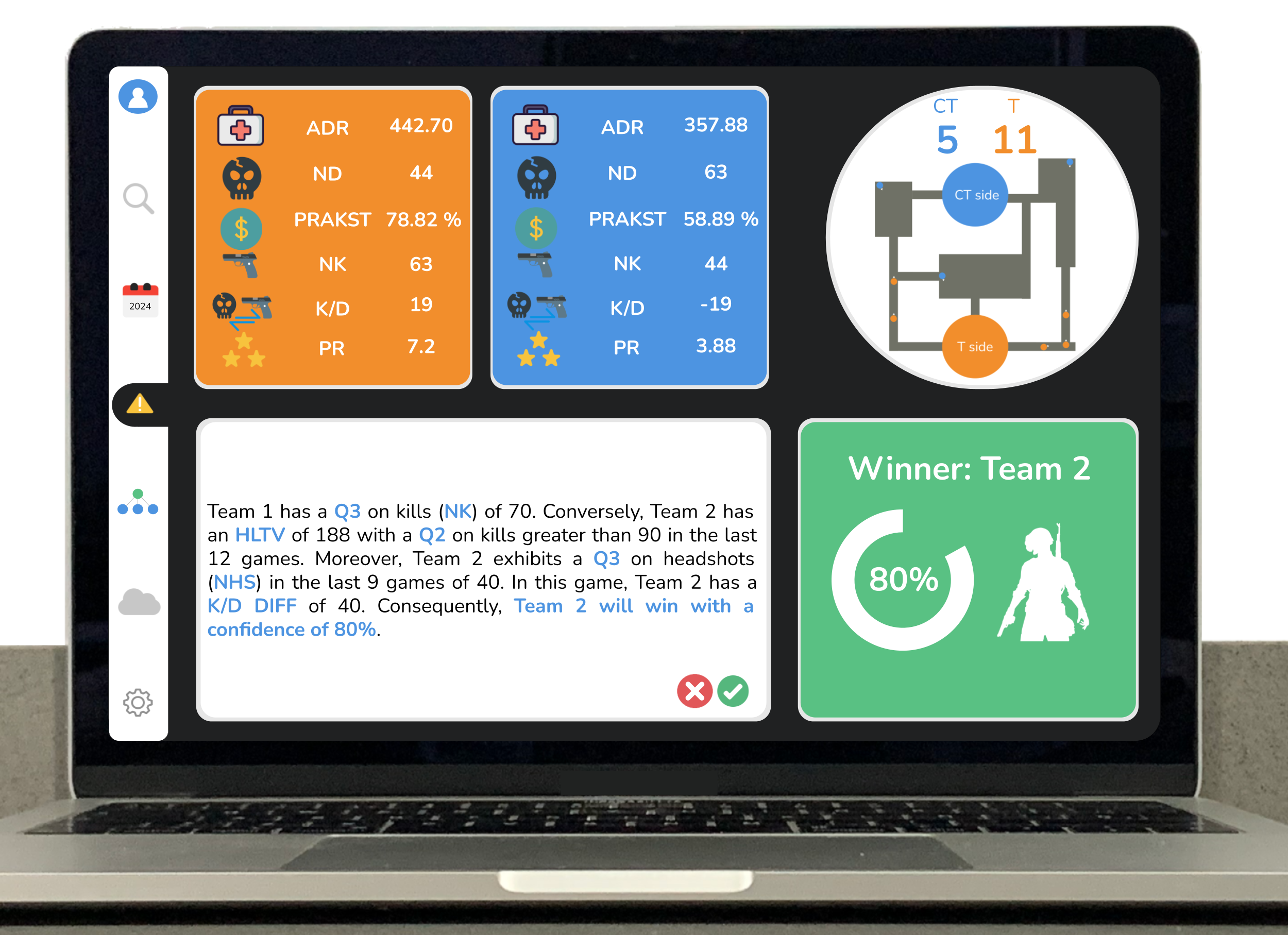}
		\caption{Explainability dashboard exploiting post-game features and combining visualization and natural language.}
		\label{fig:dasboard}
	\end{figure}

	Figure \ref{fig:dasboard} illustrates the explainability dashboard that summarizes the most relevant post-game features from the pre-game and in-game ones. 16-games 5-players accumulated feature values regarding the basic skills of the players are displayed at the top. \textsc{prakst} is calculated considering the 5 players. On the right, the scoreboard and map are displayed, showing the players' positions in real time. At the bottom, the most relevant features involved in the decision path of the \textsc{arfc} model are included in a natural language description of the prediction, with its confidence. The latter is visually summarized in the bottom right part.
	
\begin{figure}
		\centering
		\includegraphics[scale=0.12]{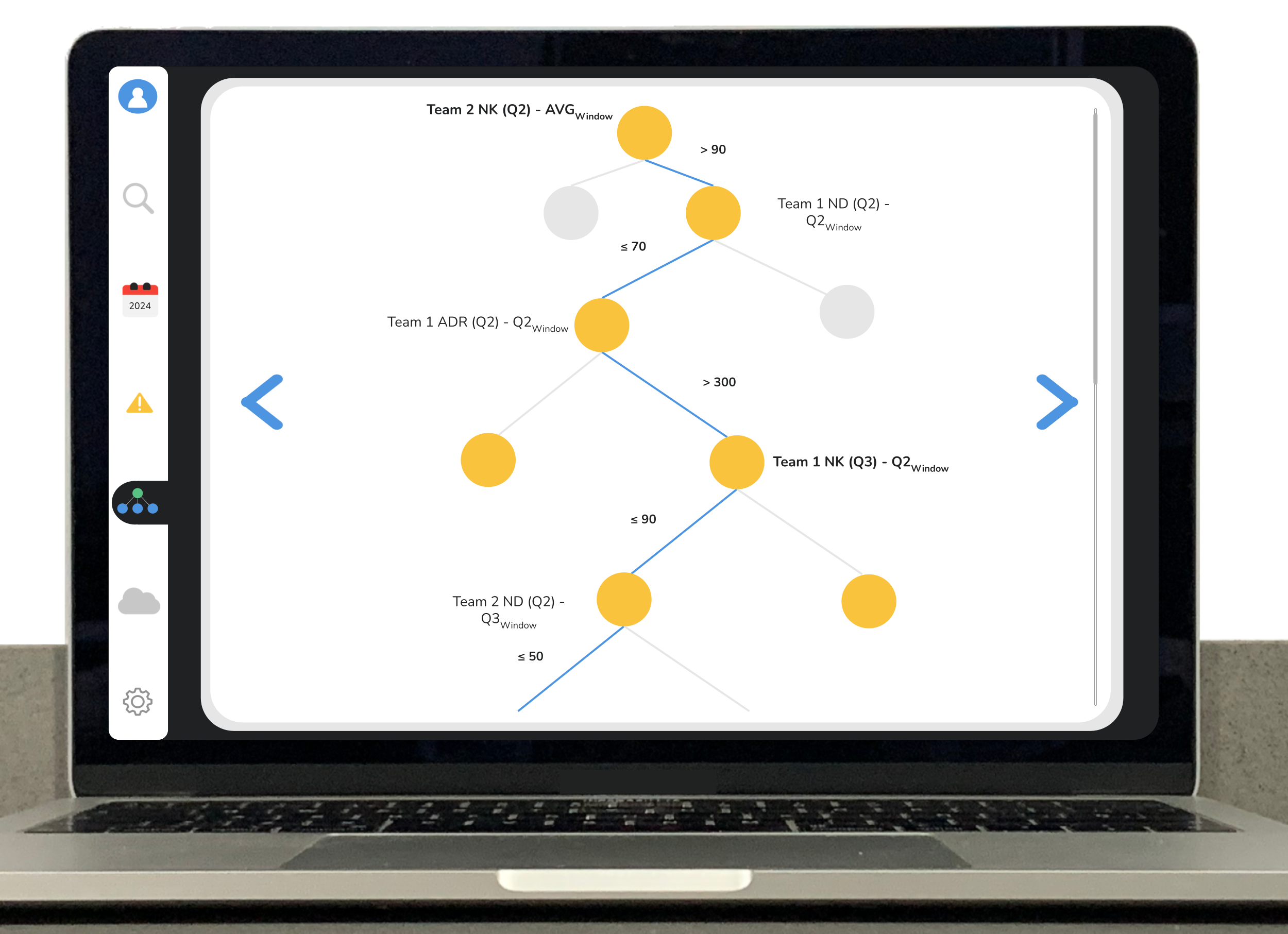}
		\caption{Decision path displayed in the explainability dashboard.}
		\label{fig:dashboard_path}
	\end{figure}

	Moreover, the dashboard allows the display of the decision paths by the tree model (see Figure \ref{fig:dashboard_path}). To ensure a user-friendly interface, the unexplored tree nodes are not shown, and those in the natural language description are in bold. Similarly, the decision tree paths that do not align with the majority-voted classification output are also removed and, thus, can not be displayed by navigating with the left and right blue arrows.
	
\begin{figure}
		\centering
		\includegraphics[scale=0.12]{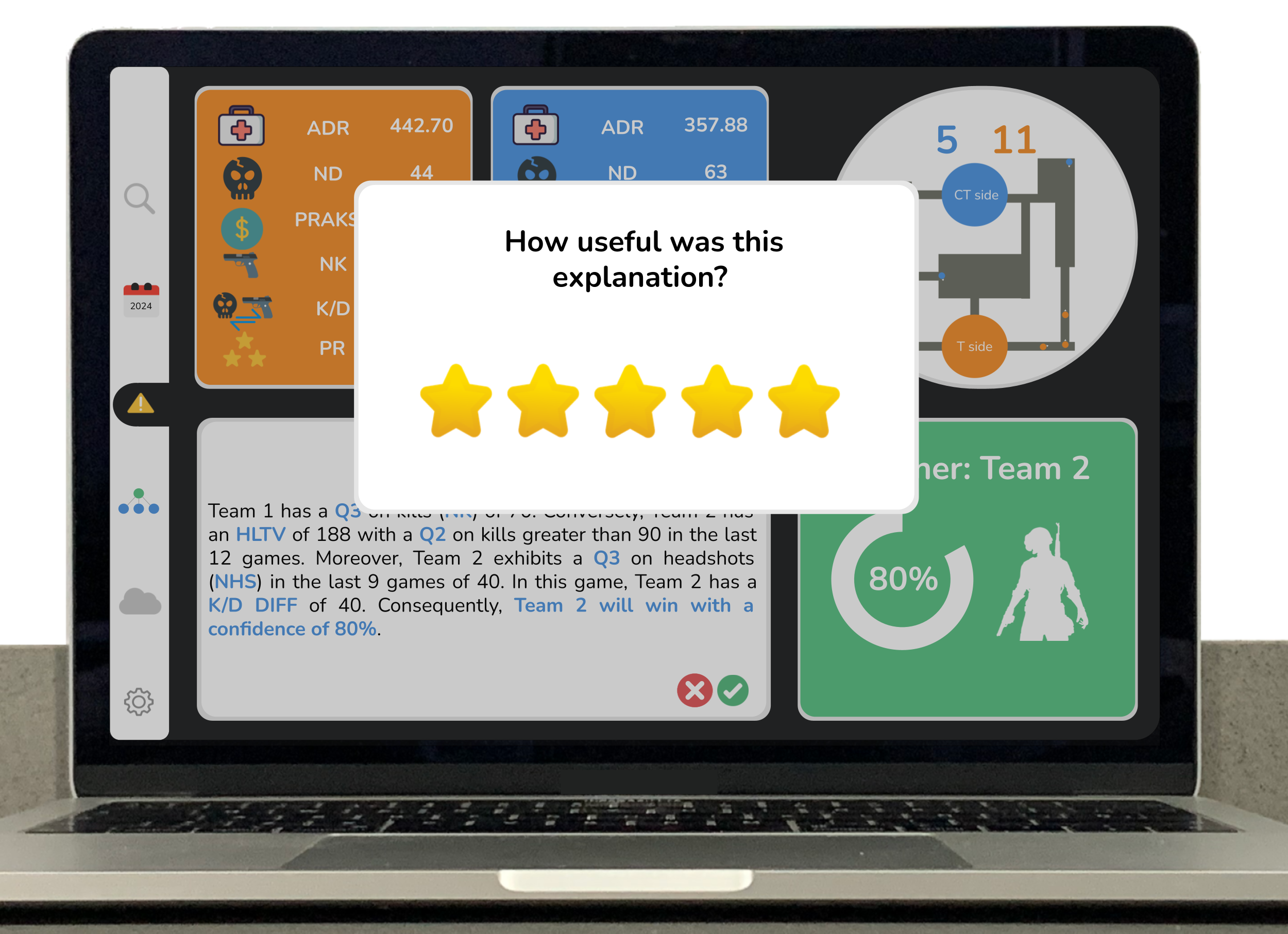}
		\caption{Explainability assessment.}
		\label{fig:dashboard_stars}
	\end{figure}

	To evaluate the quality and usefulness of the explanations provided, the end user can assess them (see Figure \ref{fig:dashboard_stars}). By clicking on the red cross, the human-in-the-loop indicates that the explanation is inadequate, while when the green tick is clicked, a 5-level Likert scale is shown for a more detailed evaluation.
	
	\section{Conclusions}
	\label{sec:conclusions}
	
	Due to the massive and rapid expansion of the gaming industry, e-sports analytics plays a vital role in academia and industry. Accordingly, most e-sports, particularly regarding the \textsc{moba}, attract millions of players and, thus, produce a publicly available stream of data regarding matches and competitions. The vast amount of data produced can be easily stored and retrieved by researchers to perform e-sports analytics, providing valuable tactical knowledge to players. In this sense, \textsc{ml} is an appropriate approach for e-sports analytics due to the availability of high-dimensional and high-volume data from \textsc{moba} e-sports. However, the lack of interpretability and explainability is one of the most significant limitations of previous works focusing on pre-match prediction, which cannot utilize live data streams as input data for the models.
	
	In light of the above, this work contributes an explainable win prediction classification solution for streaming, where input data is controlled over several sliding windows to reflect relevant game changes. Our solution addresses the research gap of jointly considering online analytics and providing explainability in win prediction. Experimental results attained an accuracy higher than \SI{90}{\percent} for win prediction in streaming.
	
	Thanks to the explainability module, the system can be exploited by ranking and recommender systems for decision-making. This module promotes trust in the outcome predictions. Moreover, to increase the value of our proposal, the system was designed to minimize the training data needed (half of the game).
	
	In future work, we plan to exploit live \textsc{api} information as input data to train our model and analyze how new features influence winning prediction (\textit{e.g.}, physiological features of players). Specifically, \textsc{api}s like Faceit\footnote{Available at \url{https://docs.faceit.com}, June 2025.}, Leetify\footnote{Available at \url{https://leetify.com}, June 2025.}, Scope.gg\footnote{Available at \url{https://scope.gg}, June 2025.} and Steam\footnote{Available at \url{https://steamcommunity.com/dev}, June 2025.} can be used. Moreover, the system will be made available to the research community via \textsc{api} to analyze user feedback, notably regarding the explainability dashboard. Furthermore, new classification tasks (\textit{e.g.}, death prediction) and perspectives (\textit{e.g.}, unsupervised classification, reinforcement learning) will be taken into account. Finally, we will examine specific aspects of the solution, such as computing and storage load, to deploy it into production.
	
	\section*{CRediT authorship contribution statement}
	
	\textbf{Silvia García-Méndez}: Conceptualization, Methodology, Software, Validation, Formal analysis, Investigation, Resources, Data Curation, Writing - Original Draft, Writing - Review \& Editing, Visualization, Project administration, Funding acquisition. \textbf{Francisco de Arriba-Pérez}: Conceptualization, Methodology, Software, Validation, Formal analysis, Investigation, Resources, Data Curation, Writing - Original Draft, Writing - Review \& Editing, Visualization, Project administration, Funding acquisition.
	
	\section*{Declaration of competing interest}
	
	The authors have no competing interests to declare relevant to this article's content.
	
	\section*{Data availability}
	
	The used data is openly available\footnoteref{data_link}.
	
	\section*{Acknowledgments}
	
	This work was partially supported by (\textit{i}) Xunta de Galicia grants ED481B-2022-093 and ED481D 2024/014, Spain; and (\textit{ii}) University of Vigo/CISUG for open access charge.
	
	\bibliography{2_bibliography}
	
\end{document}